\pdfoutput=1

\documentclass[10pt]{article}
\usepackage[explicit]{titlesec}
\usepackage{hyphenat}
\usepackage{ragged2e}
\usepackage{xcolor}

\usepackage[T1]{fontenc}
\usepackage{amsmath, amsthm, amsfonts}
\usepackage{graphicx}
\usepackage{caption}
\usepackage{subcaption}

\usepackage{soul}
\setul{.6pt}{.4pt}

\usepackage{url} 

\usepackage[ruled, linesnumbered]{algorithm2e}

\usepackage{geometry}
\usepackage{fancyhdr}
\usepackage[labelfont={footnotesize,bf} , textfont=footnotesize]{caption}
\makeatletter
\renewcommand\tagform@[1]{\maketag@@@ {\ignorespaces {\footnotesize{\textbf{Equation}}} #1.\unskip \@@italiccorr }}
\makeatother
\titleformat{\section}
  {\normalfont\fontsize{12}{15}}{\thesection}{0em}{\MakeUppercase{\textbf{#1}}}
\renewcommand{\thesection}{}
\titlespacing\section{0pt}{0pt}{-10pt}
\titleformat{\subsection}
  {\normalfont}{\thesubsection}{1em}{{\textbf{\textit{#1}}}}
\titleformat{\subsubsection}
  {\normalfont}{\thesubsection}{1em}{{\textit{#1}}}
\titlespacing{\subsubsection}{0pt}{0pt}{-10pt}
\renewcommand{\thesubsection}{\alph{subsection}}
\titlespacing\subsection{0pt}{0pt}{-8pt}

\makeatletter
\newcommand\sixteen{\@setfontsize\sixteen{16pt}{6}}
\renewcommand{\maketitle}{\bgroup\setlength{\parindent}{0pt}
\begin{flushleft}
\vspace{-.375in}
\sixteen\bfseries \@title
\medskip
\end{flushleft}
\textit{\@author}
\egroup}
\makeatother

\usepackage{multicol}
\usepackage{enumitem}
\usepackage{float}
\usepackage{arydshln}

\title{Partitioned Graph Convolution Using Adversarial and Regression Networks for Road Travel Speed Prediction}

\author{Jakob Meldgaard Kjær - jkjar14@student.aau.dk \\Lasse Kristensen - lkri15@student.aau.dk \\Mads Alberg Christensen - mach15@student.aau.dk \\\\
Supervisor: \\
Bin Yang - byang@cs.aau.dk \\\\
Department of Computer Science, Aalborg University, Aalborg, Denmark.
}

\begin{document}
\maketitle
\vspace{.12 in}

\begin{multicols}{2}

\section{Abstract} \label{sec:abstract}
Access to quality travel time information for roads in a road network has become increasingly important with the rising demand for real-time travel time estimation for paths within road networks. In the context of the Danish road network (DRN) dataset used in this paper, the data coverage is sparse and skewed towards arterial roads, with a coverage of 23.88\% across 850,980 road segments, which makes travel time estimation difficult. Existing solutions for graph-based data processing often neglect the size of the graph, which is an apparent problem for road networks with a large amount of connected road segments. To this end, we propose a framework for predicting road segment travel speed histograms for dataless edges, based on a latent representation generated by an adversarially regularized convolutional network. We apply a partitioning algorithm to divide the graph into dense subgraphs, and then train a model for each subgraph to predict speed histograms for the nodes. The framework achieves an accuracy of 71.5\% intersection and 78.5\% correlation on predicting travel speed histograms using the DRN dataset. Furthermore, experiments show that partitioning the dataset into clusters increases the performance of the framework. Specifically, partitioning the road network dataset into 100 clusters, with approximately 500 road segments in each cluster, achieves a better performance than when using 10 and 20 clusters.

\section{Introduction} \label{cha:introduction}
Graph convolutional networks (GCN) have seen a rise in importance within the transportation domain, with the increasing demand for real-time travel time estimation for paths within road networks. Analysis of traffic data from road networks is important for a variety of applications, such as path navigation \cite{zhang2018deeptravel}, traffic forecasting \cite{li2017diffusion}, and ride-sharing services \cite{riley2020realtime}. GCNs utilize strategies that convolute a graph into a low-dimensional feature space that preserves the structure of the graph data, such as edge features, vertex features, and the topology of the graph. However, a problem that often arises for GCNs in the transportation domain, is that the models generally work well on smaller graphs, but decrease drastically in either efficiency or precision when handling larger networks. For example, GCNs utilizing an embedding strategy based on the adjacency matrix of the graph, such as Kipf \& Welling \cite{variational_autoencoder}, suffer from time-expensive multiplication with larger matrices, with a per-vertex time complexity of $\mathcal{O}(n^{2.376})$ \cite{matrix_multi_complexity}. On the contrary, neural networks utilizing probabilistic models, such as node2vec \cite{node2vec} or DeepWalk \cite{deepwalk}, are more scalable to larger graphs, but suffer in accuracy compared to the adjacency models \cite{variational_autoencoder}, as the probabilistic models only capture the similarities in global graph structure and local neighborhood for each vertex, and not the features of the nodes.

These problems are especially apparent for graphs based on road networks, as the amount of road segments in these networks can be in the millions. For example, the network used in this paper, the Danish road network, has 850,980 unique road segments \footnote{According to OpenStreetMap.org, October 2019.}, with a data coverage of 23.88\%, which is a low amount of road segments compared to road networks of larger countries. With the amount of nodes in this graph, having access to a computationally efficient embedding strategy is important.
Furthermore, as each road segment in the network contains multiple features that can be used to further increase the accuracy of the embedding, having an embedding strategy that utilizes these extra node features is important for the precision of the embedding.

Recent work in graph-based neural networks have studied ways to increase the quality of the latent representation of the graph. For example, one paper introduces an adversarial training scheme to the autoencoder structure, to learn a better latent representation \cite{arga}. Another paper presents an approach that uses inductive learning to embed the nodes of a graph \cite{graphsage}.
Instead of using the traditional embedding approach of training a distinct vector for each node, this paper trains aggregator functions that are able to aggregate the node features from a node's local neighborhood. On the contrary, other work in graph-based neural networks have studied how to handle the computational downsides that occur when training larger graphs. For example, one paper introduces a batching strategy that first partitions the graph, and then stochastically combines these partitions into a number of sets with the same amount of partitions, to form sub-graphs with reduced cross-batch variance \cite{cluster-gcn}. They then use each of the sub-graphs as a batch to perform a stochastic gradient descent update. This lowers the amount of memory required to train the model, which enables training of, in theory, arbitrarily large graphs.

In this paper, we propose a novel framework that incorporates theory from both areas of the current research. The framework is inspired by the embedding strategy and adversarial training scheme from Pan et al. \cite{arga} and the stochastic batching approach introduced by Chiang et al. \cite{cluster-gcn}. Following our earlier work \cite{prob_approach_TTD}, the framework is able to embed a large road network in clusters, used for prediction of a histogram of travel speed for a specific road segment within the network, for the purposes of combating data sparseness in road network travel data. Our contributions can be summarized as follows:

\vspace{-14pt}
\begin{itemize}[leftmargin=*, topsep=0pt, nolistsep]
    \item We propose a novel GCN framework for dynamic prediction of travel speed histograms for road segments in a large road network. Our framework is able to predict travel speed histograms based on sets of adversarially regularized embeddings, each set learned from a dense subgraph of the road network.
    \item We propose a node2vec-based model for static prediction of travel speed histograms for road segments in a road network. This model is able to construct embeddings that retain feature information for road segments in a road network, and predict travel speed histograms by using them as input to a regression model.
    \item We achieve results of 71.5\% intersection and 78.5\% correlation on a road network dataset consisting of approximately 50,000 nodes and 100,000 edges.
\end{itemize}

\section{Related Work} \label{cha:related_work}
\subsection*{Models for Graph Embedding}
Graph embedding refers to the process of learning a representation of the nodes and/or edges of a graph, often resulting in a lower-dimensional representation, with the goal of retaining as much information about the graph as possible. These models have shown their applicability in a variety of application domains, such as social networks \cite{deepwalk}, image denoising \cite{image_denoising_cite}, anomaly detection \cite{anomaly_detect_cite}, and road network analysis \cite{jepsen2018network}.

Random-walk based models, such as Grover \& Leskovec \cite{node2vec} and Perozzi et al. \cite{deepwalk}, expand upon the Word2vec skip-gram model for natural language processing by Mikolov et al. \cite{word2vec}, to adhere to representation learning of graphs. These models only consider the topology of the graph as information, and the objective of the models is to retain as much information about the topology as possible. As a result of this, additional features, such as node context, will not be taken into consideration when constructing the latent representation of the graph. The work by Mikolov et al. \cite{word2vec} has also been extended to the transportation domain. Liu et al. \cite{road2vec} introduce Road2Vec, a neural network that learns road segment embeddings that capture the road interaction in urban road systems. Here, two road segments that frequently co-occour in travel routes should have a similar vector representation, similar to how two words that frequently co-occur in a sentence have similar representations in Word2vec \cite{word2vec}.

Jepsen et al. \cite{jepsen2018network} investigate whether or not existing network embedding methods are suitable for analysis tasks of a road network where only the network structure is available, specifically predicting the speed limit and road category of the roads. With only the network structure available, they rely on a high presence of homophily, which results in them using the model by Grover \& Leskovec \cite{node2vec} as the embedding technique.

With the work of Kipf \& Welling \cite{variational_autoencoder}, autoencoders have gained popularity in addressing the problem of learning a robust, lower-dimensional representation of a graph. Here, convolutional layers are used to reduce the dimensionality of an input adjacency matrix representing the graph. Thereafter, the matrix is reconstructed using an inner product decoder between the latent variables. Additionally, the autoencoder structure allows node content to be incorporated in the latent representation. Expanding on this, Pan et al. \cite{arga} propose ARGA, a framework for graph embedding using the Kipf \& Welling \cite{variational_autoencoder} autoencoder and an adversarial regularizer. Here, the regularizer uses an adversarial training scheme to force the latent representation to match a distribution, such as a Gaussian distribution. The adversarial training scheme is a neural network in itself, with its own weights, and the goal is for this model to be unable to distinguish between the latent representation and the values sampled from the distribution.

\begin{figure*}
    \centering
    \includegraphics[width=0.8\textwidth]{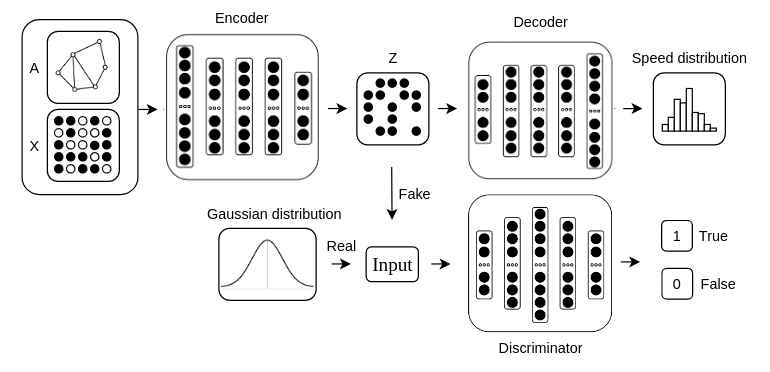}
    \caption[]{Framework architecture for Partitioned Graph Convolution Using Adversarial and Regression Networks for Road Travel Speed Prediction. The topmost component is the graph convolutional network, where the encoder encodes an adjacency matrix $A$ and node feature matrix $X$ into a latent representation $Z$. The decoder decodes the latent representation with the purpose of predicting travel speed histograms. The bottom component is the adversarial network that is used to force the latent representation to match a Gaussian distribution.}
    \label{fig:framework}
\end{figure*}

\subsection*{Models for Convolution on Large Graphs}
Recently, a branch of research in graph convolutional networks has focused on ways to allow larger graphs to be embedded, by improving on the computation time and/or memory usage of the models used to embed the graph. Since graph convolutional layers use the interaction between nodes to embed the graph, the loss term in a GCN is dependent on a large amount of nodes. Specifically, to be able to utilize back propagation in GCNs, it requires storing the intermediate embeddings for all model layers in memory. The research can be divided in two categories; sampling-based and clustering-based models.

Research has been done in the area of selective sampling, i.e. reducing, or specifying, the area in which neighbors are sampled to embed a node. Hamilton et al. \cite{graphsage} propose a model that restricts the neighborhood size for each layer, by using Monte Carlo approximation to decide on neighborhood samples of a fixed size, to be used during back-propagation. Similarly, Chen et al. \cite{fastgcn} use Monte Carlo approaches and importance sampling to reduce the size of the sampled neighborhood. Chen et al. \cite{vr-gcn} propose a model that uses a variance reduction technique to reduce the size of the neighborhood, leading to a reduction in computation time while retaining the quality of the embedding, even when reducing the neighborhood to two samples per node.

Instead of following the approach of handling larger graphs by reducing the size of the sampled neighbor nodes, Chiang et al. \cite{cluster-gcn} propose a GCN model that utilizes clustering algorithms to exploit the clusters appearing in a graph. The model partitions the input graph into dense subgraphs, and then uses the subgraph to calculate the stochastic gradient descent for the batch, reducing the amount of neighbor nodes needed to embed a node.

\section{Problem Definition and Framework} \label{cha:problem_definition}
A road network graph is represented as $H = \{I, R\}$, where $I$ is a set of nodes representing intersections in a road network, and $R$ is a set of road segments between intersections. The line graph of graph $H$ is represented as $G = \{V, E, X\}$, where $V = \{v_1, \cdots, v_n\}$ represents the road segments, i.e. the edges, of graph $H$, $E$ is a set of edges representing an intersection between two road segments, and $X \in \mathbb{R}^{N \times F}$ is a set of features for each road segment $v_i \in V$ where $N$ is the number of nodes and $F$ is the number of features in $G$. The topology of line graph $G$ is represented by an adjacency matrix $A \in \mathbb{R}^{N \times N}$, where $A_{i, j} = 1$ if a car can drive directly from the first road segment to the second, i.e. if a car can go from $v_i$ to $v_j$ via an intersection. The features of each node $v_i$ are represented by $x_i \in X$. A road segment travel speed histogram is a $k$-tuple $Hist = (b_1, \cdots, b_k)$, where each bucket $b_i$ represents the proportion of vehicles travelling at a specific speed interval on road segment $v_i$.

Given a large graph $G$, our purpose is to predict a travel speed histogram for the nodes without data in $G$, based on a low-dimensional representation of the graph. With an adjacency matrix $A$ and feature matrix $X$ as input, the low-dimensional representation should preserve information that allows for accurate prediction of the histograms.


\subsection*{Framework Overview}
The objective of this research is to predict a travel speed histogram for the nodes in a large graph $G = \{V, E, X\}$, by partitioning the graph into $s$ subgraphs $\hat{G} = [G_1, \cdots, G_s]$ and embedding each subgraph separately. The latent representation of each subgraph $G_s$ is trained to predict travel speed histograms for the nodes in $G_s$. The framework introduced in this paper is inspired by two existing studies: Pan et al. \cite{arga} and Chiang et al. \cite{cluster-gcn}. Figure \ref{fig:framework} shows an overview of the model architecture, where each subgraph $G_s \in \hat{G}$, represented by an adjacency matrix $A$ and feature matrix $X$, is used as input to its own model, giving $s$ trained models for $s$ subgraphs.

The framework consists of three distinct modules: the graph partitioning algorithm, the adversarially regularized convolutional network for convoluting the nodes of each subgraph to a set of low-dimensional vectors, and a regression model to train the vector embeddings to predict traversal speed histograms. Additionally, we introduce a random-walk based road network embedding strategy that utilizes the node2vec model from Grover \& Leskovec \cite{node2vec} to capture topology and features. This model is used as a way to compare the graph convolution model to a model that uses the probabilistic approach of random-walk. It should be noted that, even though the output of both the convolution-based and node2vec-based model is similar, i.e. travel speed histograms for road segments in a road network, the way that they achieve this result is quite different. The node2vec-based strategy is static in the way that it handles new nodes, as it embeds the entire road network at the same time so that any road segment can be directly compared to another from the start, which means that introducing a new node, such as when a new road has been introduced to the road network, will require re-embedding the entire road network. On the contrary, the convolution-based model is more dynamic in handling new nodes, as the model only embeds a node when predicting a histogram for the node. This means that new nodes will not require a complete re-embedding of the road network. The three modules of the framework are as follows.

\vspace{-14pt}
\begin{itemize}[leftmargin=*, topsep=0pt, nolistsep]
    \item \textbf{Graph Partitioning}. The graph partitioning algorithm takes graph $G$ as input, alongside the amount of partitions and batches wanted, and outputs a set of tuples containing an adjacency matrix $A_i$ and feature matrix $X_i$ for each partitioned subgraph $G_i$.
    
    \item \textbf{GCN with Adversarial Network}. The graph convolutional network takes one tuple from the graph partitioner as input, i.e. an adjacency matrix $A_i$ and feature matrix $X_i$, and convolutes the graph into a lower-dimensional latent representation $Z_i$. The network also includes a regression model that predicts a travel speed histogram for a road segment, by decoding the latent representation of the segment to the dimensions of the histogram. Additionally, an adversarial network takes the latent representation and forces it to suit a prior distribution using an adversarial training scheme. The purpose of the adversarial network is to distinguish whether a latent vector $z \in Z_i$ is from the prior distribution or the encoder.
    
    \item \textbf{node2vec Feature Inclusion Model}. The node2vec feature inclusion model takes a weighted directed graph $G$ as input, and generates a set of node sequences from the graph using random-walk. The model then uses Word2vec \cite{word2vec} to vectorize the graph nodes based on the sequences. In addition to the topology, the sequences are used to embed the features of the node.
\end{itemize}


\section{Methodology} \label{cha:methodology}

\subsection{Graph Partitioning}
\label{sec:clustering}
Inspired by Chiang et al. \cite{cluster-gcn}, the graph partitioning algorithm aims to partition graph $G = \{V, E, X\}$ into $s$ dense subgraphs: $\hat{G} = [\{V_1, E_1, X_1\}, \cdots, \{V_s, E_s, X_s\}]$. The dense subgraphs are identified using the METIS \cite{metis} partitioning algorithm. METIS aims to partition the graph so that within-cluster edges are more prevalent than between-cluster edges, leading to the identification of the dense clusters of the graph. Intuitively, this also means that the neighborhood of a given node is most likely included within the partition that the node belongs to.

Additionally, we also employ the stochastic multiple partitions approach from Chiang et al. \cite{cluster-gcn}. Instead of only considering one partition as a batch for a stochastic gradient descent update, we stochastically combine $q$ partitions together to form a single batch $V = \{V_1 \cap \cdots \cap V_q\}$. In addition to each partition's nodes and edges being included in the batch, the between-cluster edges, i.e. the edges between partitions that were part of the original graph, are re-added to the graph. The approach of having multiple partitions in a single batch reduces the across-batch variance, and leads to an improved convergence rate \cite{cluster-gcn}.

\subsection{GCN with Adversarial Network}
\label{sec:arga}

The GCN module draws inspiration from Pan et al. \cite{arga} and their ARGA framework. Given an adjacency matrix $A$ and feature matrix $X$, the encoder produces an embedding matrix $Z \in \mathbb{R}^{N \times M}$, where $M$ is the embedding dimension, with each row vector $z_i \in Z$ representing a vertex in the graph. The architecture of the decoder depends on whether the dataset is road data or the Cora dataset. For road data, the decoder predicts a speed histogram for each vertex, and as such the output of the decoder is a matrix $Y \in \mathbb{R}^{N \times k}$, where $k$ is the amount of buckets in the speed histogram. For classification on the Cora dataset, the output of the decoder is $Y \in \mathbb{R}^{N \times C}$, where $C$ is the amount of paper topics, i.e. labels. The encoder and decoder are accompanied by a discriminator that aims to distinguish between embeddings created by the encoder and random numbers drawn from a normal distribution, producing a discrimination matrix $B \in \mathbb{R}^{N \times 1}$. This is done to force the encoder to create embeddings that match a normal distribution in order to reduce overfitting.

The encoder consists of two graph convolution layers with a Gaussian noise layer in between. Each graph convolution layer computes
$$H^{(l+1)} = \text{ReLU} \left( \hat{D}^{-\frac{1}{2}} \hat{A} \hat{D}^{-\frac{1}{2}} H^{(l)} W^{(l)} \right)$$
where $\text{ReLU} = \max(0, x)$, $\hat{A} \in \mathbb{R}^{N \times N}$ is the adjacency matrix with an added self-connection for each vertex (such that $\hat{A} = A+I_N$, where $I_N$ is the $N$-dimensional identity matrix), $\hat{D} \in \mathbb{R}^{N \times N}$ is the degree matrix from $\hat{A}$, $H^{(l)} \in \mathbb{R}^{N \times O}$ with $ H^{(0)} = X$ is the output of layer $l$, $W^{(l)}$ is the weight matrix of layer $l$, and $O$ is the amount of output dimensions for layer $l$.
The discriminator is a three-layer model, where each layer computes
$$H^{(l+1)} = f\left( H^{(l)} W^{(l)} \right)$$

where \[
f = \begin{cases}
        \text{ReLU}\left( x \right)&\text{if $l \in \{0,1\}$}\\
        \frac{1}{1 + e^{-x}} &\text{if $l = 2$}
    \end{cases}
\]

The entire model is trained in a supervised manner. The pseudocode for each training step is outlined in algorithm \ref{algo:optimization}. For the algorithm, $Q$ is random data drawn from the standard normal distribution, $S$ and $T$ are histograms with $k$ buckets each, $\nabla$ is the gradient of a loss function with respect to model weights, $\text{mean}$ is the mean of all matrix elements, $\max$ is the element-wise maximum operation, $\odot$ is the element-wise matrix multiplication operation, $J_{N,1}$ is a column-vector of ones, and $0_{N,1}$ is a column vector of zeros.

If the model is used on road data, the decoder loss function is histogram intersection between the predictions and labels, as seen on line 7 in algorithm \ref{algo:optimization}, while it is categorical cross-entropy for the Cora dataset, as seen on line 5.
On lines 8 and 9, the binary cross entropy loss for the discriminator's prediction for encoder embeddings compared to $0_{N,1}$, as well as the discriminator's prediction for $Q$ compared to $J_{N,1}$, is calculated.

On line 15 in algorithm \ref{algo:optimization}, the binary cross-entropy loss is calculated for the embeddings created by the encoder as well as the predictions by the discriminator.
Weights for the encoder, decoder, and discriminator are updated according to the loss gradients.

\IncMargin{1em}
\begin{algorithm}[H]
\SetAlgoLined

\Indm
\SetKwInOut{Input}{Input}
\SetKwInOut{Output}{Output}
\Input{Feature matrix $X \in \mathbb{R}^{N \times F}$\\Adjacency matrix $A \in \mathbb{R}^{N \times N}$\\Label matrix $Y \in \mathbb{R}^{N \times C}$ (Cora) OR\\$Y \in \mathbb{R}^{N \times k}$ (road data)}

\Indp
$Q \in \mathbb{R}^{N \times O} \sim \mathcal{N}\left( \mu=0, \sigma^2=1 \right)$\;
$Z = encoder.\textsc{call}\left(X, A\right)$\;
$U = decoder.\textsc{call}\left(Z\right)$\;
\BlankLine

\uIf{$\text{dataset is Cora}$}{$L_1 = -\frac{1}{N} \sum^{N}_{i=1} \log P\left[u_i \in C_{u_i}\right]$\;}
\uElse{$L_1 = 1 - \sum^{k}_{i=1} \min \left( S_i, T_i \right)$\;}
Update $encoder$ and $decoder$ weights given $\nabla L_1$\;
\BlankLine

$R = discriminator.\textsc{call}\left( Q \right)$\;
$F = discriminator.\textsc{call}\left( Z \right)$\;
$L_{2_R} = \text{mean}\left(\max\left( R, 0 \right) - R \odot J_{N, 1} + \log\left(1 + \rm{e}^{-\lvert R \rvert} \right)\right)$\;
$L_{2_F} = \text{mean}\left( \max\left( F, 0 \right) - F \odot 0_{N, 1} + \log\left(1 + \rm{e}^{-\lvert F \rvert} \right)\right)$\;
$L_2 = L_{2_R} + L_{2_F}$\;
Update $discriminator$ weights given $\nabla L_2$\;

\BlankLine
$L_3 = \text{mean} \left( \max\left( F, 0 \right) - F \odot J_{N, 1} + \log\left(1 + \rm{e}^{-\lvert F \rvert} \right) \right)$\;
Update $encoder$ weights given $\nabla L_3$\;

\caption{Optimization step}\label{algo:optimization}
\end{algorithm}
\DecMargin{1em}

\subsection{node2vec Feature Inclusion Model}
\label{sec:node2vec_comparison_model}
The node2vec feature inclusion model is a modification of the random-walk based framework of node2vec \cite{node2vec}. node2vec is a graph-based extension of the Word2Vec skip-gram neural network, and works by utilizing random-walk on a graph to generate sequences of nodes to be used as input to the Word2Vec model. Word2Vec then uses an unsupervised skip-gram model to embed each node into a vector of a given number of dimensions. node2vec is, in general, used to embed the topology of a graph, as node context is not considered in the skip-gram model. The node2vec feature inclusion model entertains the idea of embedding features separately to introduce them to the model. Essentially, this model embeds the topology and each feature separately, and then concatenates them into a single vector. The combined vector is then used as input to a regression model that predicts a travel speed distribution based on the vectors. 

\begin{table*}[!ht]
    \centering
    \begin{tabular}{|l|l|c|c|c|c|}
        \hline
        Dataset & Task & \#Nodes & \#Edges & \#Features \\
        \hline
        Cora & Node Classification & 2,708 & 5,429 & 1,433 \\ 
        DRN $\geq 50$ & Regression & 49,544 & 98,646 & 14 \\ \hline
    \end{tabular}
    \caption{Dataset Statistics}
    \label{table:datasets}
\end{table*}

\begin{figure}[H]
    \centering
    \includegraphics[width=0.45\textwidth]{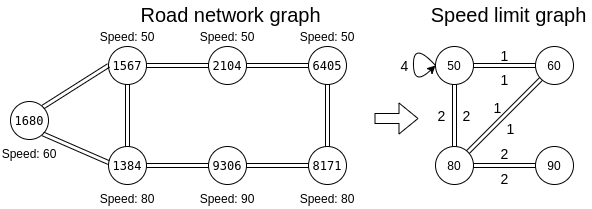}
    \caption[]{Illustration of node2vec using weigthed feature graphs}
    \label{fig:Node2Vec_v1}
\end{figure}

We introduce two different approaches for embedding the features of the graph; one that minimizes the size of the graph using weights, and one that retains the topology of the road network for the features. We will henceforth refer to the first approach as the feature graph approach, and the second as the sequence manipulation approach. For the feature graph approach, we first do random-walk on the road network graph to generate the node sequences for the topology of the road network. Thereafter, we construct a weighted directed graph for each of the $n$ features $\hat{G} = [G_1, \cdots, G_n] = [\{V_1, E_1\}, \cdots, \{V_n, E_n\}]$. The graphs are constructed as visualized in figure \ref{fig:Node2Vec_v1}. An edge between two speed limits is added and/or weighted if an edge between two road segments, i.e. nodes in the road network graph, with the two speed limits exists. We then perform random-walk on each feature graph, generating a set of sequences for each feature. We then embed each sequence, i.e. the topology and feature sequences, and concatenate the embeddings to form a single vector embedding for each node in the road network.

\begin{figure}[H]
    \centering
    \includegraphics[width=0.45\textwidth]{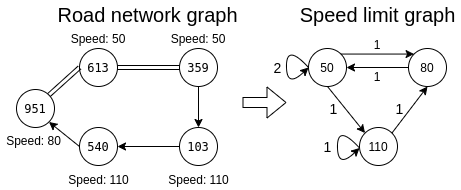}
    \caption[]{Weighted feature graph with incorrect sequencing}
    \label{fig:Node2Vec_approach1_example}
\end{figure}
The sequence manipulation approach exploits the topological road network graph to create a set of feature sequences to be used as input in the Word2Vec model. For this approach, we again first generate sequences based on the topological road network graph, i.e. the leftmost graph in figure \ref{fig:Node2Vec_v1}, and embed it. Generating sequences for the topology could, as an example, give the following sequences of road segment identifiers, using a sequence length of five:
\begin{align*}
    S_1 = [1680, 1384, 1567, 2104, 6405]\\
    S_2 = [9306, 1384, 1680, 1567, 2104]
\end{align*}
Now, for each feature, we replace the road segment identifier with its feature value. For example, replacing with the speed limit of the road segment will give the following sequences.
\begin{align*}
    S_1 = [60, 80, 50, 50, 50]\\
    S_2 = [90, 80, 60, 50, 50]
\end{align*}
We then use the constructed feature sequences to embed each feature, and then concatenate the topology and each feature embedding to form a single vector for each node in the road network. Doing this, we ensure that the generated sequences always represent the actual feature sequences from the road network, ensuring that we will not embed a feature using a sequence that is incorrect. This will not always be the case for the feature graph approach. For example, figure \ref{fig:Node2Vec_approach1_example} shows a directed road network graph and its associated weighted directed speed limit graph. Using this graph, the sequence $S = [50, 110, 80, 50, 110]$ can be generated by using a random-walk process on the speed limit graph. But, according to the road network graph, it is not possible to generate a sequence with a speed limit of $[50, 110, 80]$. Therefore, conducting random-walk on the topological road network graph, and then replacing the road segment identifiers in the sequences with their feature values, will ensure that only sequences that adhere to the topology of the graph are generated, resulting in feature embeddings that are more representative of real-world scenarios.
\begin{table}[H]
    \centering
    \begin{tabular}{|l|c|c|c|c|}
        \hline
        Model & Int. & Cor. & Bhat. & KL div. \\
        \hline
        Base node2vec & 0.443 & 0.486 & 0.531 & 4.010 \\
        Feature Graph & 0.705 & 0.637 & 0.367 & 2.103 \\
        Sequence Manip. & \textbf{0.741} & \textbf{0.671} & \textbf{0.326} & \textbf{1.472} \\
        \hline
    \end{tabular}
    \caption{Comparison of node2vec approaches}
    \label{table:node2vec_approaches}
\end{table}
This is also further solidified by a comparison of the accuracy of the two approaches. Table \ref{table:node2vec_approaches} shows a comparison of using the embedding for prediction, for the base node2vec algorithm on the topology and the two feature-inclusion approaches. As can be seen, sequence manipulation beats the feature graph approach across all four metrics, with a significant improvement in KL divergence. Furthermore, there is a significant improvement to both approaches compared to base node2vec, suggesting that the inclusion of the feature embeddings increases the accuracy of prediction using the generated embeddings. We will in the experiments be using the sequence manipulation approach as a baseline.

\section{Experiments} \label{cha:experiments}
We evaluate the performance of our framework on two datasets with two different tasks; the Cora \cite{cora} dataset and the Danish road network (DRN) introduced in \cite{prob_approach_TTD}. The dataset statistics can be seen in table \ref{table:datasets}. Furthermore, we evaluate the effectiveness of applying the clustering strategy when using the DRN dataset, in terms of accuracy vs. execution time. For the DRN dataset, the amount of travel speed observations necessary for a node to be considered is set to $\geq 50$, with a total amount of travel speed observations across road segments of 41,734,487. The model and experiments are implemented using TensorFlow 2.2.0 \cite{tensorflow} with the included Keras API \cite{keras}. For all experiments, the travel speed histogram consists of 22 buckets $Hist = (b_1, \cdots, b_{22})$, with a bucket spread of $2 m/s$. This makes the top bucket contain values of $42 m/s \leq i < 44 m/s$, which means that we maintain speeds up to $\approx 158 km/h$.

\begin{table*}[!ht]
    \centering
    \begin{tabular}{|l|c|c|c|c|c|c|c|c|}
        \hline
         \textbf{DRN}    & \multicolumn{4}{c}{Mean} \vline & \multicolumn{4}{c}{Median} \vline\\ \cline{2-9} 
        \textbf{Regression} & Int. & Cor. & Bhat. & KL div. & Int. & Cor. & Bhat. & KL div. \\
        \hline
        Naive Baseline 1    & 0.153 & 0.370 & 0.603 & 1.474 & 0.154 & 0.401 & 0.599 & 1.433 \\
        Naive Baseline 2    & 0.203 & 0.528 & 0.467 & \textbf{0.900} & 0.207 & 0.588 & 0.448 & 0.809 \\
        Base N2V            & 0.443 & 0.486 & 0.531 & 4.010 & 0.444 & 0.518 & 0.530 & 3.591 \\
        N2V Features        & 0.665 & 0.734 & 0.335 & 1.736 & 0.714 & 0.851 & 0.292 & 0.571 \\ 
        ARGA TF2            & 0.517 & 0.580 & 0.467 & 3.258 & 0.532 & 0.646 & 0.454 & 2.317 \\ \hdashline
        GCN w/o adversarial & 0.699 & 0.767 & 0.330 & 1.991 & 0.772 & 0.904 & 0.270 & 0.507 \\
        Full GCN framework  & \textbf{0.715} & \textbf{0.785} & \textbf{0.314} & 1.436 & \textbf{0.773} & \textbf{0.907} & \textbf{0.266} & \textbf{0.446} \\
        \hline
    \end{tabular}
    \caption{Prediction Results on DRN}
    \label{table:DRN_experiment}
\end{table*}

\subsection{Prediction on DRN Dataset} \label{experiments:regression}
We compare our proposed GCN framework with the node2vec-based approaches and a reimplementation of ARGA \cite{arga}, in terms of accuracy on travel speed histogram prediction for the DRN dataset. All experiments have been conducted using the same hardware.

\textbf{Baselines.} We include the following models in the DRN prediction experiment. All node2vec-based models include a regression model that takes the embeddings produced by node2vec as input and predicts a histogram of travel speed for each node embedding. The regression model is an MLP with two fully-connected layers, where the first layer has 32 output neurons and ReLU for activation, and the second layer has 7 output neurons and softmax for activation. For the ARGA model, the experiments have been conducted on a dense subgraph of the dataset, consisting of approximately 5,000 road segments, due to limitations regarding memory usage. For both the full GCN framework and the GCN without adversarial, the experiments have been conducted on a partitioned dataset of 100 batches.
\vspace{-12pt}
\begin{itemize}[leftmargin=*, topsep=0pt, nolistsep]
    \item Naive Baseline 1: No learning. Instead, compute the average travel speed histogram of all road segments with histogram data in the graph, and use it as the prediction for the road segments without data.
    
    \item Naive Baseline 2: No learning. Similar to Naive Baseline 1, but computing the average travel speed histogram for each speed limit.
    
    \item Base N2V: The base node2vec algorithm, utilizing random-walk to generate sequences of nodes to be used in the Word2Vec skip-gram model. Only the topology is embedded.
    
    \item N2V Features: node2vec with sequence manipulation for feature embeddings, i.e. our second proposed approach for the node2vec feature inclusion model.
    
    \item ARGA TF2: Re-implementation of ARGA \cite{arga} in Tensorflow 2.2.0, with the addition of a regression model to predict travel speed histograms based on the embeddings produced by ARGA.
    
    \item GCN w/o adversarial: Our proposed GCN framework without the adversarial training scheme.
    
    \item Full GCN framework: The partitioned adversarial GCN for histogram prediction. Our full proposed framework for prediction of road segment travel speed histograms.
\end{itemize}

\textbf{Metrics.} The results of prediction on the DRN dataset are reported in terms of Intersection, Correlation, Bhattacharyya distance, and Kullback-Leibler (KL) divergence, between a predicted histogram of travel speed and the real histogram. For intersection and correlation, the higher the value the better, and for Bhattacharyya distance and KL divergence, the lower the better. Each experiment has been conducted 10 times and we report the mean and median values of each metric. We use $2/3$ of the dataset for training and $1/3$ for testing. For the full GCN framework, we use $2/3$ of each batch for training and $1/3$ for testing.

\textbf{Parameter Settings.} All models have been trained for 2000 epochs, and are optimized using the Adam optimizer. Through a parameter study, the learning rate of the GCN has been set to $10^{-3}$ and the discriminator learning rate has been set to $10^{-4}$. Furthermore, the encoder of the GCN has a hidden layer with 32 neurons and an embedding layer with 16 neurons, and the decoder has two hidden layers with 256 neurons and an output layer with 22 neurons. The decoder and the discriminator has a dropout rate of 30\%. The discriminator has two layers with 64 and 32 neurons respectively, and an output layer with one neuron. For the node2vec-based models, the default parameters for the reference node2vec implementation have been used.

\textbf{Results.} The experimental results on prediction of histograms on the DRN dataset are shown in table \ref{table:DRN_experiment}. It shows that the full GCN framework proposed in this paper performs best across all metrics except for the KL divergence mean, where Naive Baseline 2 performs the best. Furthermore, the median results show that the majority of the predictions are above 90\% in correlation and 77\% in intersection, with the KL divergence being significantly better compared to the mean.  Regarding the adversarial training scheme, the addition of this shows an increase in performance for the mean values of each metric, but not a notable increase for the median. This indicates that predictions of a lower accuracy, i.e. the outliers in the dataset, have increased in performance for each metric with the addition of the adversarial training scheme.

\begin{table*}[!ht]
    \centering
    \begin{tabular}{|l|c|c|c|c|c|c|}
        \hline
        \textbf{Cora} & \multicolumn{3}{c}{Mean} \vline & \multicolumn{3}{c}{Median} \vline \\ \cline{2-7}
        \textbf{Classification} & Accuracy & F1 & ROC AUC & Accuracy & F1 & ROC AUC \\
        \hline
        Base N2V              & 0.690 & 0.673 & 0.905 & 0.684 & 0.673 & \textbf{0.912} \\
        ARGA TF2              & 0.662 & 0.649 & \textbf{0.906} & 0.665 & 0.647 & 0.911 \\ \hdashline
        GCN w/o adversarial   & \textbf{0.707} & \textbf{0.681} & \textbf{0.906} & \textbf{0.704} & \textbf{0.682} & 0.905 \\
        GCN w/o partitioning  & 0.653 & 0.613 & 0.887 & 0.663 & 0.623 & 0.893 \\
        GCN 5 batches         & 0.422 & 0.295 & 0.719 & 0.456 & 0.302 & 0.686 \\
        GCN 20 batches        & 0.269 & 0.134 & 0.577 & 0.296 & 0.136 & 0.572 \\
        \hline
    \end{tabular}
    \caption{Classification results on Cora}
    \label{table:Cora_experiment}
\end{table*}

\subsection{Node Classification} \label{experiments:classification}
\textbf{Baselines.} We include the following models in the Cora classification experiment. Base N2V is node2Vec with default parameters, as well as an MLP with two fully-connected layers. The first layer in the MLP has 32 output neurons and ReLU for activation, and the second layer has 7 output neurons and softmax for activation. All experiments have been conducted using the same hardware.
\vspace{-13pt}
\begin{itemize}[leftmargin=*, topsep=0pt, nolistsep]
    \item Base N2V: The base node2vec algorithm, utilizing random-walk to generate sequences of nodes to be used in the Word2Vec skip-gram model. Only the topology is embedded.
    
    \item ARGA TF2: Re-implementation of ARGA \cite{arga} in Tensorflow 2.2.0, with the addition of a regression model to predict travel speed histograms based on the embeddings produced by ARGA.
    
    \item GCN w/o adversarial: Our framework without the adversarial training scheme. The dataset is a single cluster, i.e. no partitioning.
    
    \item GCN w/o partitioning: Our proposed framework for prediction of road segment travel speed histograms, without partitioning the dataset.
    
    \item GCN 5 batches: Our proposed framework, with the dataset being partitioned in 5 batches.
    
    \item GCN 20 batches: Our proposed framework, with the dataset being partitioned in 20 batches.
\end{itemize}

\textbf{Metrics.} The results of classification on the Cora dataset are reported in terms of Accuracy, Macro-F1 score, and Area Under the Receiver Operating Characteristic Curve (ROC AUC) score. Categorical crossentropy has been used as the loss function for the regression model in these experiments. Each experiment has been conducted 10 times and we report the mean and median values for each metric. We use the training, test, and validation split introduced in \cite{cora_split}.

\textbf{Parameter Settings.} All models have been trained for 2000 epochs, and are optimized using the Adam optimizer. The dataset has been split randomly, but still such that there are 20 nodes for each of the seven classes for training, 30 for validation, and the rest for training.  For the node2vec-based models, the default parameters for the reference node2vec implementation. Through a parameter study, the learning rate of the GCN has been set to $10^{-4}$ and the discriminator learning rate has been set to $10^{-5}$. Furthermore, the encoder of the GCN has a hidden layer with 32 neurons and an embedding layer with 32 neurons, and the decoder has two hidden layers with 16 neurons and an output layer with 7 neurons. The decoder has a dropout rate of 20\%. The discriminator has two layers with 64 and 32 neurons respectively, an output layer with one neuron, and a dropout rate of 50\%. The activation functions for each of the final layers of the encoder, decoder, and discriminator are linear, softmax, and linear, respectively. Every other layer has ReLU as activation function.

\textbf{Results.} Table \ref{table:Cora_experiment} shows the results of the Cora classification experiment. The results show that the our proposed GCN framework without the adversarial training scheme performs the best across all metrics except ROC AUC. This might indicate that, when the task at hand is to classify a node, and when the dataset consists of few features, the adversarial training scheme only adds noise to the model. As the purpose of the adversarial training scheme is to regularize the latent representation, when simple data is handled this only weakens the model. Furthermore, increasing the number of batches from 1 to 5 to 20 lead to a decrease in performance across all metrics. This might indicate that the tendencies of the dataset has to be fully represented in each batch, and, in our case, the smaller batches generated by our partitioning algorithm is not ideal for representing the Cora dataset.

\begin{table*}[!ht]
    \centering
    \begin{tabular}{|l|c|c|c|c|c|c|}
        \hline
         \textbf{DRN}    & \multicolumn{4}{c}{Mean} \vline & Computation & Memory \\ \cline{2-5} 
        \textbf{Clustering} & Int. & Cor. & Bhat. & KL div. & time & usage \\
        \hline
        (10, 10)  & 0.647 $\pm$ 0.004 & 0.716 $\pm$ 0.006 & 0.367 $\pm$ 0.004 & 1.832 $\pm$ 0.077 & 00:52:31.233 & 3.02 GB \\
        (40, 10)  & 0.647 $\pm$ 0.002 & 0.713 $\pm$ 0.003 & 0.367 $\pm$ 0.002 & 1.821 $\pm$ 0.046 & 00:51:44.747 & 2.94 GB \\
        (20, 20)  & 0.673 $\pm$ 0.003 & 0.742 $\pm$ 0.003 & 0.348 $\pm$ 0.003 & 1.682 $\pm$ 0.059 & 00:35:09.294 & 2.19 GB \\
        (200, 20) & 0.673 $\pm$ 0.002 & 0.741 $\pm$ 0.003 & 0.348 $\pm$ 0.003 & 1.678 $\pm$ 0.037 & 00:36:00.298 & 2.10 GB \\
        (100, 100) & \textbf{0.715} $\pm$ 0.003 & \textbf{0.785} $\pm$ 0.003 & \textbf{0.314} $\pm$ 0.002 & \textbf{1.436} $\pm$ 0.037 & 00:45:23.527 & 1.74 GB \\
        \hline
    \end{tabular}
    \caption{Effects of Clustering when Predicting on DRN}
    \label{table:clustering_experiment}
\end{table*}

\subsection{Effects of Graph Partitioning} \label{experiments:partitioning}
We experiment on whether or not using the graph partitioning module has an effect on the computation time and accuracy of the model, specifically for prediction of travel speed histograms on the DRN dataset. As we are not able to predict on the entire dataset at the same time, we will here include experiments on having a lower vs. larger amount of clusters, and the impact of stochastic multiple partitions on accuracy. All experiments have been conducted using the same hardware.

\textbf{Baselines.} All experiments on the effects of graph partitioning have been conducted using the full GCN framework proposed in this paper. The graph partitioning module has been utilized to create a number of batches, each trained using its own model. We train each batch setting once and compute the mean and standard error of the mean for each experiment.

\textbf{Metrics.} The results are reported in terms of Intersection, Correlation, Bhattacharyya distance, and Kullback-Leibler (KL) divergence. For intersection and correlation, the higher the value the better, and for Bhattacharyya distance and KL divergence, the lower the better. Additionally, we report the computation time and memory usage of each model. The numbers in the left column of table \ref{table:clustering_experiment} represent the amount of clusters generated by the partitioning module, and the amount of batches created. For example, $(40, 10)$ indicates that the dataset has been partitioned in 40 clusters, which have been combined to form 10 batches, i.e. 4 clusters per batch. If clusters is equal to batches, e.g. $(20, 20)$, stochastic multiple partitions have not been utilized as there is one cluster per batch. The number of models trained is equal to the number of batches, and we use 2/3 of each batch for training and 1/3 for testing.

\textbf{Parameter Settings.} All models have been trained for 2000 epochs, and are optimized using the Adam optimizer. We are using identical parameter settings for the GCN as the setup used for the DRN experiment seen in table \ref{table:DRN_experiment}.

\textbf{Results.} The results in table \ref{table:clustering_experiment} show that, when increasing the number of batches to be trained, and thus reducing the size of each batch, the computation time is reduced significantly while the accuracy across the four metrics is increased. What should be noted here is that the computation time represents the time that the models have been trained sequentially, while it is fully possible to train the models in parallel, which makes the per-batch training time an important factor to study. 

For the (10, 10) and (40, 10) settings, the average training time per batch is approximately 5 minutes and 15 seconds, whereas for the (20, 20) and (200, 20) settings it is approximately 1 minute and 45 seconds on the same hardware. This is a significant speed-up in training time, while also achieving better results on all four metrics and reducing the memory usage. Again, increasing the number of batches to 100, i.e. the (100, 100) setting, we still see a significant increase in performance across all four metrics, while further reducing the per-batch training time to approximately 27 seconds. The performance increase when there are more batches indicates that larger batches is not ideal when predicting on the DRN dataset. The results also indicate that combining clusters to form batches, using the stochastic multiple partitions approach, does not increase the performance of the model. For both experiments where this has been applied, i.e. comparing (10, 10) to (40, 10) and comparing (20, 20) to (200, 20), the performance of the model has not increased, whereas the computation time has slightly increased due to the overhead from the stochastic multiple partitions algorithm.



\section{Conclusion} \label{cha:conclusion}
In this paper, we proposed a novel graph convolution framework for prediction of travel speed histograms for road segments within a road network. We argue that most graph convolutional networks used in the transportation domain suffer from time- or memory-expensive computation when handling larger road networks. We proposed a clustered graph convolutional network that learns to predict travel speed histograms based on the latent representation, by training a model for each dense cluster identified using a partitioning algorithm. The network learns jointly with an adversarial training scheme that forces the latent representation to match a Gaussian distribution. Additionally, we propose a node2vec-based model that is able to embed both the topology and features of a graph, by concatenating separate embeddings for the topology and each feature. Experiments show that the GCN framework performs best on the DRN dataset across all metrics when compared to the baselines, achieving an accuracy of 71.5\% intersection and 78.5\% correlation. Furthermore, results show that partitioning the road network into smaller batches increases the performance of the framework, while reducing the memory usage and per-batch computation time.


 \bibliography{bibtex/litterature}

\end{multicols}

\end{document}